\documentclass{article}

\usepackage[preprint]{spconfa4}
\usepackage{amssymb,amsmath}
\usepackage{framed,multirow}
\usepackage{latexsym}
\usepackage{array}
\usepackage{balance}
\usepackage{booktabs}
\usepackage{caption}
\usepackage{color}
\usepackage{graphicx}
\usepackage{multirow}
\usepackage{times}
\usepackage{url}
\usepackage{threeparttable}
\usepackage{float}

\makeatletter
\newif\if@restonecol
\makeatother

\usepackage[linesnumbered,ruled,vlined]{algorithm2e}%[ruled,vlined]
\usepackage{algpseudocode}
\usepackage{amsmath}
\newcommand\etal{\emph{et~al.}}
\newcommand\ie{\emph{i.e.}}

\begin{document}
\UseRawInputEncoding

\title{Human-In-The-Loop Document Layout Analysis}

\name{Xingjiao Wu$^{1,2,^{\dagger}}$\thanks{$^{\dagger}$ These authors contributed equally to this work.}, Tianlong Ma$^{1,2,^{\dagger}}$, Xin Li$^{2}$, Qin Chen$^{2}$, Liang He$^{1,2,*}$\thanks{$^*$Corresponding author.}}
\address{\small{1 Shanghai Key Laboratory of Multidimensional Information Processing, East China Normal University, Shanghai, China}\\
\small{2 School of Computer Science and Technology, East China Normal University, Shanghai, China}\\
{\small wuxingjiao2885@gmail.com, tlma@cs.ecnu.edu.cn, 51194506020@stu.ecnu.edu.cn, qchen@ica.stc.sh.cn, lhe@cs.ecnu.edu.cn}
}

\maketitle

\begin{abstract}
Document layout analysis (DLA) aims to divide a document image into different types of regions. DLA plays an important role in the document content understanding and information extraction systems.
Exploring a method that can use less data for effective training contributes to the development of DLA.
We consider a Human-in-the-loop (HITL) collaborative intelligence in the DLA.
Our approach was inspired by the fact that the HITL push the model to learn from the unknown problems by adding a small amount of data based on knowledge.
The HITL select key samples by using confidence. However, using confidence to find key samples is not suitable for DLA tasks.
We propose the Key Samples Selection (KSS) method to find key samples in high-level tasks (semantic segmentation) more accurately through agent collaboration, effectively reducing costs.
Once selected, these key samples are passed to human beings for active labeling, then the model will be updated with the labeled samples.
Hence, we revisited the learning system from reinforcement learning and designed a sample-based agent update strategy, which effectively improves the agent's ability to accept new samples.
It achieves significant improvement results in two benchmarks (DSSE-200 (from 77.1\% to 86.3\%) and CS-150 (from 88.0\% to 95.6\%)) by using 10\% of labeled data.

\end{abstract}


\section{Introduction}
\label{sec:intro}
The Document Layout Analysis (DLA) is an important task dedicated to extracting semantic information from the document image.
As a critical preprocessing step of document understanding systems, DLA can provide information for several applications such as document retrieval, content categorization, and text recognition.
With the development of deep convolutional neural networks, the high-capacity supervised learning algorithms have achieved remarkable results in DLA tasks~\cite{binmakhashen2019document, he2017multi, wick2018fully, li2018deeplayout, li2021few, lu2021probabilistic}.
The scale of these models has included hundreds of millions of parameters. This paradigm allows the model to have more degrees of freedom enough to awe-inspiring description power. However, the large number of parameters requires a massive amount of training data with labels~\cite{zhou2014learning}.
Improving model performance by data annotation has two crucial challenges.
On the one hand, the data growth rate is far behind the growth rate of model parameters, so data growth has primarily hindered the further development of the model.
On the other hand, the emergence of new tasks has far exceeded the speed of data updates. Suppose you have developed an industrial model for defect detection.
If the production line needs to adjust short-term production goals, you have to label new data and train the model.
The time of data annotation will become the bottleneck of the application.

Many tasks generate samples to construct new datasets, thereby speeding up model iteration and reducing the cost of data annotation~\cite{yu2015lsun,li2020layoutgan,liu2019learning,zhao2020differentiable}.
However, the generated data cannot fully satisfy the intricate details in the task, and the quality of the generated data is difficult to be effectively guaranteed.
For DLA tasks, some samples in the different datasets are similar. If we can accurately find and label the key samples, our model can learn Knowledge better and reduce data annotation pressure.
The central issue addressed in this paper is the following:
\emph{How we find the key samples on the new dataset that do not exist in the original dataset?}

\begin{figure}
	\centering
	\includegraphics[width=1\linewidth]{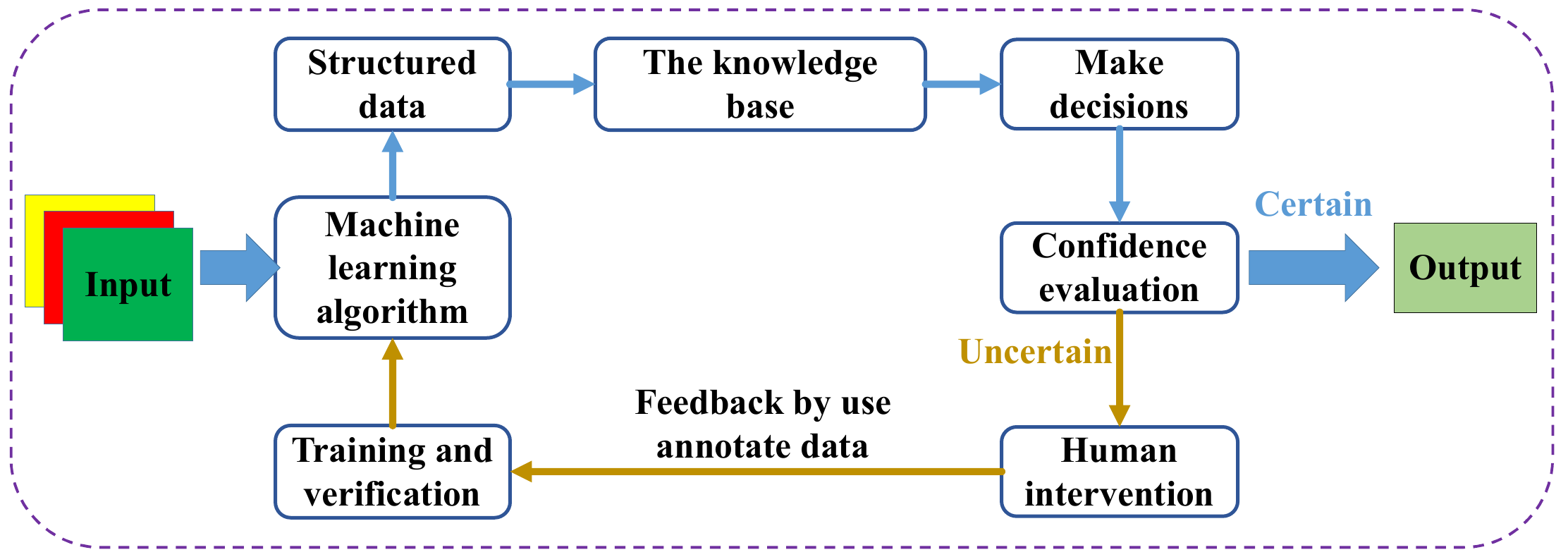}
	\caption{A Human-in-the-loop Machine Learning Pipeline. }
	\label{Fig0}
\end{figure}

To tackle this challenge, the Human-in-the-loop (HITL) methods are proposed~\cite{wan2020human, yue2020interventional, zhang2019invest}, and the pipeline shown in Fig.~\ref{Fig0}.
HITL uses each sample's confidence to judge whether the key samples.
The use of confidence to find key samples is a milestone in the classification task. However, confidence cannot be directly obtained in some tasks, the confidence's magic is not attractive. For example, the segmentation task is a pixel-level classification, confidence cannot be used to find the key samples, so we need to explore a way to find the key samples.
When people explore unknown problems, they usually need necessary discussions.
First, different people will make necessary inferences based on their knowledge; then put forward their thinking about this problem; finally, they discuss and use the discussion's consensus to define this problem.
So, we propose a framework for multi-agent collaboration under the framework of HITL.

This framework's motivation is to consider that when people learn about unknown problems, they always reason based on existing knowledge and only need a few key pieces of information to update knowledge.
Our framework consists of three parts, \textbf{the agent initialization, the agent collaboration, and the agent update}.
(1) Agent initialization: the agent should know more scenes and things, but it is unnecessary for detailed information in the domain.
This process may be similar to most models' current training using ImageNet, which means we need to use more data to train our initial model. These data do not require a perfect fit with the target task. Maybe it is related domain data or machine-synthesized data, as long as the amount of data is huge.
Inspired by data generation, we train the initial model by using generate some initial data.
We design a sample synthesis engine base on LaTex to synthesize a large amount of data as the initial knowledge carrier.
We will describe the specific synthesis process in section 3.1.
(2) Agent collaboration: agent collaboration is to discover the key samples.
Most samples in the new dataset are similar to the original dataset, the data difference only depends on a few samples.
In this process, We use two pre-trained models to deal with the unlabeled data to obtain forecast results, and then we compare the pixel classification results of the two output results. Finally, we get the samples by filter threshold.
(3) Agent update: the agent update uses human-labeled data to train the agent to learn new knowledge. Follow the reinforcement learning, and we will consider an update method that distinguishes the sample distribution.

This paper mainly makes three contributions:
\begin{itemize}
\item We propose a multi-agent collaboration framework under the Human-in-the-loop that can accurately discover key sample using
as few human efforts as possible.
To the best of our knowledge, we are the first to propose the framework for agent collaboration under the HITL.
\item We propose a key samples method that can more accurately find key samples in high-level tasks (semantic segmentation) through agent collaboration, thus effectively reducing the human efforts.
\item
We design a sample-based agent update strategy, which effectively improves the agent's ability to accept new samples.
\item We apply our framework to the DSSE-200~\cite{yang2017learning} and CS-150~\cite{clark2015looking} DLA datasets and find that it outperforms the state-of-the-art approaches.
\end{itemize}

The rest of this paper is organized as follows. Section~\ref{sec:related} introduces background of document layout analysis and Human-in-the-loop.
Section~\ref{sec:method} discusses the model design and network architecture in detail.
In Section~\ref{sec:exp}, we demonstrate the qualitative and quantitative study of the framework.
Finally, we conclude our work in Section~\ref{sec:conclusion}.

\section{Related Works}
\label{sec:related}

\subsection{Document Layout Analysis}
Document layout analysis (DLA) methods can be divided into three categories known as bottom-up, top-down, and hybrid.
The bottom-up strategy can be subdivided into five categories, and the representative work has the following:
specifically connected component analysis~\cite{tran2015hybrid}, texture analysis~\cite{mehri2017texture}, learning-based analysis, Voronoi diagram~\cite{lu2005constructing}, and Delaunay triangulation~\cite{vasilopoulos2017complex}.
The top-down strategy can be subdivided into four categories. The representative work has the following: texture-based Analysis~\cite{asi2015simplifying}, Run Length Smearing Algorithm (RLSA)~\cite{swaileh2015multi}, DLA projection-profile~\cite{shafait2010effect} and White space analysis~\cite{shafait2008background}.
The hybrid methods offer a balance between bottom-up and top-down techniques.
Chen~\etal~\cite{chen2013hybrid} propose an effective hybrid method for page segmentation, which extracts blank rectangles based on connected component analysis, and uses foreground and background information to filter blank rectangles to form new separators. Many researchers use multi-level homogeneity structure (MHS) for document layout analysis~\cite{tran2017robust,tran2016page}.
Besides, many methods based on neural networks~\cite{he2017multi, wick2018fully, li2018deeplayout, li2021few, lu2021probabilistic} have been proposed and achieved remarkable results.
Rencently, the DLA task can also be considered as a semantic segmentation task, which is to perform a pixel-level understanding of the segmentation object~\cite{ravi2016semantic,wang2017hierarchically,wu2021document}.
Xu~\etal~\cite{xu2018multi}  train a multi-task FCN to segment the document image into different regions.
Soullard~\etal~\cite{soullard2020multi} propose a fully convolutional neural network architecture (FCN) to deal with historical newspaper images by using a pixel labelling of the various semantic entities.
Zheng~\etal~\cite{zheng2019content}  propose a deep generative model for graphic design layouts to synthesize layout designs.
Zheng~\etal~\cite{li2020cross} propose a novel cross-domain DOD model to learn a detector for the target domain using labelled data from the source domain and only unlabeled data from the target domain.
Xu~\etal~\cite{xu2020layoutlm}  propose the LayoutLM to jointly model interactions between text and layout information across scanned document images.
Many of these papers employ the Fully Convolutional Network (FCN, \cite{long2015fully}) for semantic segmentation.

As pointed out by Binmakhashen~\etal~\cite{binmakhashen2019document} in their survey, the deep-learning DLA methods require a long training time and huge data.
Therefore, exploring a method that can use less data for effective training contributes to the development of document layout.

\subsection{Human-In-The-Loop}

\textbf{Active Learning }
Active learning is a commonly used technique in machine learning��which involves humans labeling the most interesting examples iteratively ~\cite{lewis1994sequential,ricci2015recommender}.
Active learning has many excellent jobs, such as Uncertainty sampling, Query-by-committee (QBC), Expected model change, Expected error reduction, and Expected error reduction~\cite{chai2020human}.
However, the agents in these algorithms cannot try to optimize external rewards, and therefore the challenges involved in combining autonomous reinforcement learning with human expertise not be tackled~\cite{mandel2017add}.
Our agent collaboration is inspired by QBC~\cite{seung1992query,grollman2007dogged,chernova2009interactive,judah2011active,silver2012active,judah2014active}.
However, there are two differences between KSS and QBC. One is that our labeled samples use a data generation method that does not require manual labeling, and the other is that our agents try to optimize external rewards.
Besides, as far as we know, this is the first work to introduce agent collaboration into HITL.

\textbf{Human-In-The-Loop }
With the development of convolutional neural networks, the growth rate of model capacity has far exceeded the available data development scale.
To enable the model to quickly learn to `solve' a new task by giving it a few training examples of a new task, some few-shot learning approaches have been proposed~\cite{ren2018meta, andrychowicz2016learning}.
To allow the network to effectively learn novel categories from only a few training data while at the same time it will not forget the initial categories, Gidaris \etal~designed a few-shot visual learning system~\cite{gidaris2018dynamic}.
Yao~\etal~proposed a graph few-shot learning (GFL) algorithm, which shares a transferable metric space characterized by node embedding and graph-specific prototype embedding functions between the auxiliary graph the target, thereby promoting the transfer of structural knowledge. GFL learns prior knowledge from the auxiliary graph to improve the classification accuracy of the target graph~\cite{yao2020graph}.
The well-known network structure of meta-learning also includes Siamese neural network~\cite{koch2015siamese}, matching network~\cite{vinyals2016matching} and prototype network~\cite{snell2017prototypical}.

Many generate-based methods can also expand the size of the dataset well. In particular, the method of using Generative Adversarial Networks (GAN)~\cite{goodfellow2014generative} has achieved state-of-the-art performance on many tasks~\cite{wu2017survey}.
To make better use of the image generator's semantic layout, Liu~\etal~\cite{liu2019learning} used a convolution kernel conditioned on the semantic label map to generate an intermediate feature map from the noise map and finally generated an image.
Liu~\etal~\cite{li2020layoutgan} proposed a novel LayoutGAN to generate the layout of relational graphic elements.

HITL selects the most useful samples to improve conventional machine learning tasks' performance has become a hot topic in recent research.
Li~\etal \cite{li2017human} proposed a hybrid human-machine data integration framework for data integration problems that purely automated methods cannot completely address.
In response to the lack of data samples in current learning methods, Maxime~\etal \cite{chevalier2018babyai} used the HITL method to develop a platform named BabyAI to support investigations towards including humans in the loop for grounded language learning.
To address the challenge that deep learning needs to rely on a large amount of manually labeled data, Zhang~\etal \cite{zhang2019invest} proposed a framework that combines the advantages of regular expressions and deep learning.
Many tasks have almost no annotated data and always need to be created from scratch. To address this challenge, Klie~\etal \cite{klie2020zero} proposed a novel domain-agnostic HITL annotation approach.
Yue~\etal \cite{yue2020interventional} proposed a framework named Interventional Few-Shot Learning (IFSL) to address an overlooked deficiency in recent FSL methods.
Wan~\etal \cite{wan2020human} proposed a HILL learning algorithm to dynamically reject uncertain predictions and mark them to increase a novel set of categories.

The above methods are largely considered to deal with the lack of data, and they are all excellent work. However, these methods have certain limitations in finding error-prone samples. Therefore, we propose an agent collaboration method to select key samples.

\section{Approach}
\label{sec:method}

\begin{figure*}
	\centering
	\includegraphics[width=0.92\linewidth]{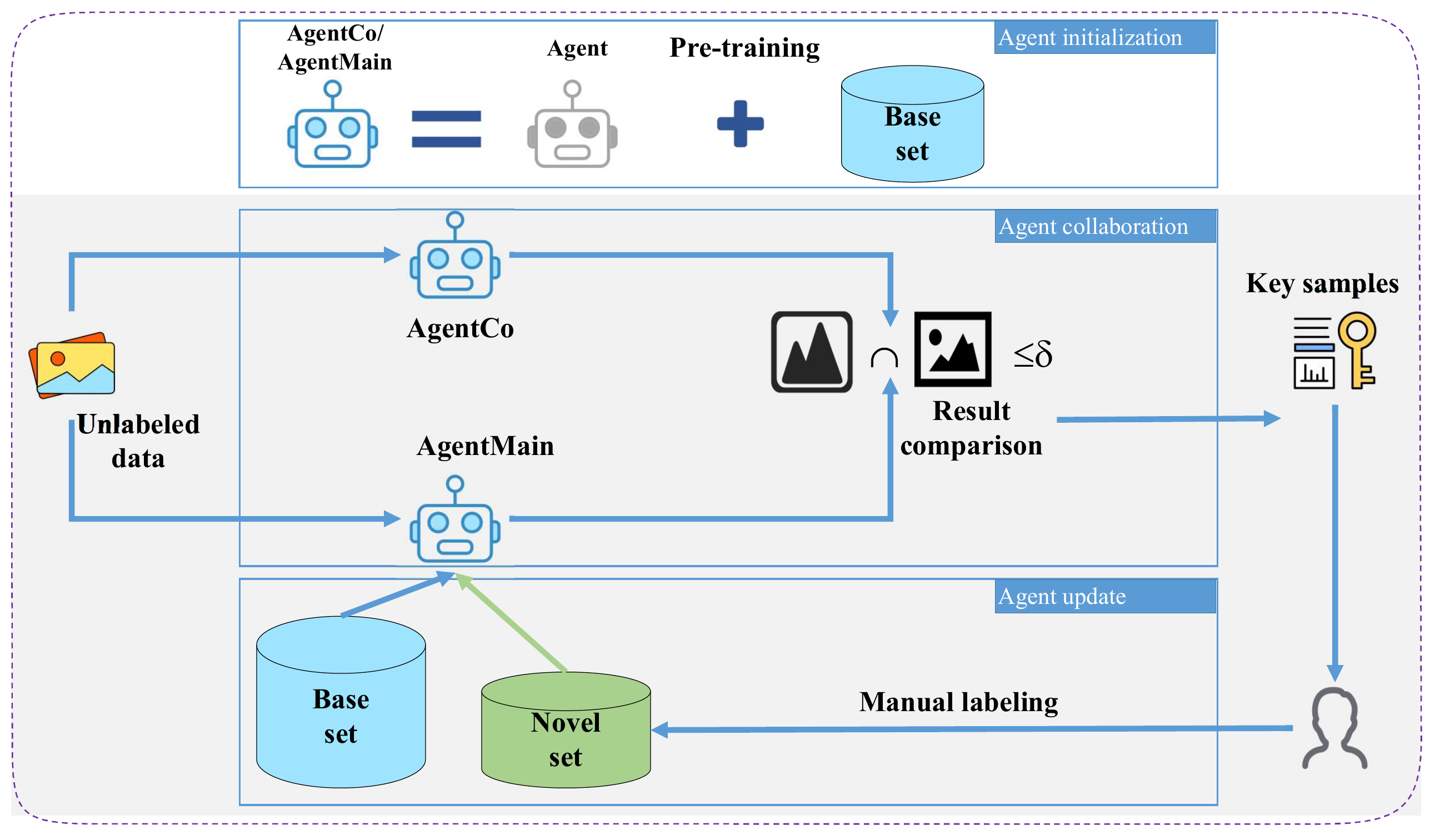}
	\caption{ The architecture of our framework. It consists of three parts: agent initialization, agent collaboration
and agent update. }
	\label{Fig1}
%\vspace{-8px}
\end{figure*}

As shown in Fig.~\ref{Fig1}, our framework consists of three parts: agent initialization, agent collaboration, and agent update.
First, we generate a large-scale sample ($D_{base}$) using a sample synthesis engine base on LaTex.
Then we use the large-scale datasets to train two general agent models.
Next, we use unlabeled samples ($D_{novel}$) as input prediction results using two agents, and we compare the two prediction results to obtain a key sample with large prediction differences.
Finally, we make manual labeling these key samples and use them as input to retrain the model.

We assume there are $K_{base}$ base samples, and there are $N_b$ categories:

\begin{equation}
{D_{base}} = \bigcup\limits_{b = 1}^{{K_{base}}} {\left\{ {{x_b},i} \right\}} _{i = 1}^{{N_b}}
\label{equ:Dbase}
\end{equation}

Where $K_{base}$ is the number of base samples, ${x_b},i$ is the $i-th$ sample in base samples b that is composed of $N_b$ class.
Similarly, we can define the novel set as Eq.~\ref{equ:Dnovel}

\begin{equation}
{D_{novel}} = \bigcup\limits_{n = 1}^{{K_{novel}}} {\left\{ {{x_n},i} \right\}} _{i = 1}^{{N_n}}
\label{equ:Dnovel}
\end{equation}

\textbf{The agent initialization }is to establish a general model. To briefly describe this process, we use $f(.)$ to represent the main agent model's training process and use $g(.)$ to represent the training process of the collaborative agent model.
we define the ${x_m} \in {\mathbb{R}^m} = f({\hat x_b})$ as the feature vector extracted from the base data ${\hat x_b} ({\hat x_b} \in {D_{base}})$ via function $f(.)$, we can get a training weight set as $W_m^* = \bigcup\nolimits_{m = 1}^{{N_b}} {\left\{ {\omega _m^*} \right\}} $.
Similarly, we define the ${x_c} \in {\mathbb{R}^m} = f({\hat x_b})$ as the feature vector extracted from the base data ${\hat x_b} ({\hat x_b} \in {D_{base}})$ via function $g(.)$,
we can get a training weight set as $W_c^* = \bigcup\nolimits_{c = 1}^{{N_b}} {\left\{ {\omega _c^*} \right\}} $.

\textbf{The agent collaboration } is to select key samples, input the new sample into the trained agent, compare the results' intersection, and select the key samples in the two agents. We can get the output of the main agent model as Eq.~\ref{equ:Ouputm}, and the output of the collaborative agent model as Eq.~\ref{equ:Ouputc}.

\begin{equation}
{O^m}({\hat x_n}) = \delta ({\tau _m} \cdot {x_m} \cdot {\omega _m} \cdot {\hat x_n})
\label{equ:Ouputm}
\end{equation}

\begin{equation}
{O^c}({{\hat x}_n}) = \delta ({\tau _c} \cdot {x_c} \cdot {\omega _c} \cdot {{\hat x}_n})
\label{equ:Ouputc}
\end{equation}

Where $\delta(.)$ is a post-processing operation, and its function is to map the final feature matrix through operations such as softmax to obtain the most likely classification result matrix. ${\hat x_b}$ from the novel data ${\hat x_n} ({\hat x_n} \in {D_{novel}})$, $\tau _m$ and $\tau _c$ are learnable parameter.
By solving the number of intersections of the two samples' results, we can help us get the key samples. Assume $\hat x$ is a matrix size is $w \times h$, and $SC(.)$ is used as a scaling factor, the size of the output matrix of the agent is $SC(w) \times SC(h)$. We can define the score of the key samples as Eq.~\ref{equ:ScoreK}.

\begin{equation}
{ScoreK({{\hat x}_n}) = {O^m}({{\hat x}_n}) \cap {O^c}({{\hat x}_n})}
\label{equ:K}
\end{equation}

\begin{equation}
ScoreK({\hat x_n}) = \frac{{\sum\limits_{i = 1}^{SC(w)} {\sum\limits_{j = 1}^{SC(h)} {(\left[ {O_{ij}^m = O_{ij}^m} \right])} } }}{{SC(w) \times SC(h)}}
\label{equ:ScoreK}
\end{equation}

\textbf{Key 1: } (Validation of Key Samples Selection) The previous HITL framework did not consider processing pixel-level classification, so there is no formed sample selection strategy. Here we set three confidence scoring strategies (Eq.~\ref{equ:con1}, Eq.~\ref{equ:con2}, Eq.~\ref{equ:con3}) based on the usual confidence selection algorithm, and cooperate with our multi-agent strategy comparison.
We assume that the final output feature is $F_{whd}$, $w$ represents the feature width at this time, $h$ represents the feature height, and $d$ represents the number of feature channel layers. At this time, $d$ and the number of classifications are the same as $N_b$.

\begin{equation}
MACON : \left\{ {\begin{array}{*{20}{c}}
{f_{ij}^1 = \frac{1}{{1 + {e^{ - \max \{ {F_{ijk}}\} _{k = 1}^d}}}}}\\
{co{n_1} = \frac{{\sum\limits_{i = 1}^w {\sum\limits_{j = 1}^h {\left( {{{f_{ij}^1} \mathord{\left/
 {\vphantom {{f_{ij}^1} {\sum {{f^1}} }}} \right.
 \kern-\nulldelimiterspace} {\sum {{f^1}} }}} \right)} } }}{{w \times h}}}
\end{array}} \right.
\label{equ:con1}
\end{equation}

\begin{equation}
SUCON : \left\{ {\begin{array}{*{20}{c}}
{f_{ij}^2 = \frac{1}{{1 + {e^{ - sum\{ {F_{ijk}}\} _{k = 1}^d}}}}}\\
{co{n_2} = \frac{{\sum\limits_{i = 1}^w {\sum\limits_{j = 1}^h {\left( {{{f_{ij}^2} \mathord{\left/
 {\vphantom {{f_{ij}^2} {\sum {{f^2}} }}} \right.
 \kern-\nulldelimiterspace} {\sum {{f^2}} }}} \right)} } }}{{w \times h}}}
\end{array}} \right.
\label{equ:con2}
\end{equation}

\begin{equation}
MECON : \left\{ {\begin{array}{*{20}{c}}
{f_{ij}^3 = \frac{1}{{1 + {e^{ - median\{ {F_{ijk}}\} _{k = 1}^d}}}}}\\
{co{n_3} = \frac{{\sum\limits_{i = 1}^w {\sum\limits_{j = 1}^h {\left( {{{f_{ij}^3} \mathord{\left/
 {\vphantom {{f_{ij}^3} {\sum {{f^3}} }}} \right.
 \kern-\nulldelimiterspace} {\sum {{f^3}} }}} \right)} } }}{{w \times h}}}
\end{array}} \right.
\label{equ:con3}
\end{equation}

We will verify the method effect in the section~\ref{sec:ABstudy}.

\textbf{The agent update } is to use new samples to train the model so that the model's ability to recognize unknown samples can be effectively enhanced.
This process can be expressed as Eq.~\ref{equ:update}.

\begin{equation}
\omega _m^* \leftarrow \gamma (D_{key})
\label{equ:update}
\end{equation}
Where $\gamma (.)$ is a retrain operation, $D_{key}$ is a mixed dataset, we will explain in detail next.

\textbf{Key 2: } (Validation of Agent Update) We try to directly use select examples for fine-tuning (Eq.~\ref{equ:update1}), use select data plus original data for fine-tuning (Eq.~\ref{equ:update2}), use mix data to retrain the model (Eq.~\ref{equ:update3}), and use reinforcement learning for training (Eq.~\ref{equ:update4}).
Since the number of select samples is very limited, we will crop new sample images, tables, and text resources to prevent the agent from falling into overfitting.
Moreover, at the same time, select some materials from the original material library and use a sample synthesis engine base on LaTex to automatically expand a part of the data ($D_{synthesis}$), and then select samples and synthetic data to form a new training sample ($D'_{novel}$). The process is shown in Eq.~\ref{equ:DN}.

\begin{equation}
{D'_{novel}} = {D_{novel}} \cup {D_{synthesis}}
\label{equ:DN}
\end{equation}

\begin{equation}
PSFT : \left\{ {\begin{array}{*{20}{c}}
{\omega _m^* \leftarrow \omega _m^* \otimes \gamma (D_{key})}\\
{(D_{key} \not\subset {D_{base}}) \cap (D_{key} \subset {D'_{novel}})}
\end{array}} \right.
\label{equ:update1}
\end{equation}

\begin{equation}
PBSFT : \left\{ {\begin{array}{*{20}{c}}
{\omega _m^* \leftarrow \omega _m^* \otimes \gamma (D_{key})}\\
{(D_{key} \subset {D_{base}}) \cup (D_{key} \subset {D'_{novel}})}
\end{array}} \right.
\label{equ:update2}
\end{equation}

\begin{equation}
RBSFT : \left\{ {\begin{array}{*{20}{c}}
{\omega _m^* \leftarrow \gamma (D_{key})}\\
{(D_{key} \subset {D_{base}}) \cup (D_{key} \subset {D'_{novel}})}
\end{array}} \right.
\label{equ:update3}
\end{equation}

\begin{equation}
RBSRE : \left\{ {\begin{array}{*{20}{c}}
{\omega _m^* \leftarrow \omega _m^* \otimes ({\lambda _1} * \gamma ({D_{hb}}) + {\lambda _2} * \gamma ({D_{hn}}))}\\
{{D_{hb}} \Rightarrow D_{key} \subset {D_{base}}}\\
{{D_{hn}} \Rightarrow D_{key} \subset {D'_{novel}}}
\end{array}} \right.
\label{equ:update4}
\end{equation}

where $ \otimes $ represent the weight update, ${{\lambda _1}}$ and ${{\lambda _2}}$ are hyperparameters, we will implement them in the next experiment.

\subsection{Agent Initialization}

The purpose of agent initialization is to establish a general model, the general model can learn the prior knowledge from the original data.
The model's performance is positively correlated with the number of parameters, and the situation that the model with more parameters can handle is more complicated.
However, more parameters mean that more samples are needed for parameter constraints.
Therefore, the prerequisite for building a general model is to have a sufficient sample to train.
There is currently very limited manual labeling data in the DLA field to the best of our knowledge, so we need to find an unlabeled method that can quickly construct data samples.
Inspired by the work of Yang~\etal~\cite{yang2017learning}, we design a sample synthesis engine base on LaTex.

The idea of the generation process is shown in Algorithm 1.
The specific process can be similar to playing a jigsaw puzzle in the document.
Different from Yang~\etal, we added more details.
The caption also includes a complex font besides different sizes.
So we have also collected a title library with screenshots besides simply changing the font size.
That will contain many unforeseen deviations using Wikipedia as a source of text, the novel texts published on the Internet as the data source to improve text quality.
Besides, our data sources are from network resources, but more of them are collected and intercepted manually by our organization, and the quality is more guaranteed.
The number of data sources is also rich enough.
In addition to using MSCOCO~\cite{lin2014microsoft} directly, we also grabbed nearly 20,000 images on the Internet and manually collected 5,000 images from the magazines and web pages.

We generated 3000 samples use the LaTex synthesis engine, and we divided the training set and the validation set with a ratio of 8: 2.
The sample layout generated is relatively simple. To not cause excessive dependence on the datasets, we limited the accuracy to below 95\% during pre-training.

\begin{algorithm}[t]
  \caption{The Exapmple Generation}
  \KwIn{$PageNumber$ : The number of document, \\
  $Set$ : The attribute of the generated document ;}
  \KwOut{$Document$ : Generated document, \\
  $label$ : The Ground-True;}

  \For {$i=1; i \le PageNumber$}{

  Randomly set the font size, randomly set the margins, and randomly set whether to double-column\\

  \While {The remaining page height is not 0}
  {
  Randomly select the generation type, and calculate the page remaining according to different types.
  } %1
  } %0

return $Document , label$\;

\end{algorithm}

\subsection{Agent Collaboration}
We trained two general-purpose agents by using the generated data. The next is to use these two trained agents to collaborate to identify key data.
We discovered the fact in the migration learning experiment that the migrated model can already handle most of the data for new datasets, so we need supplementation or human intervention only a few.
However, if all the results have to be manually filtered, labor costs will be more expensive.
We propose to use two models with different structures for identifying key data. Moreover, in section \ref{sec:ABstudy}, we will verify the effect of using two models with different structures for identifying key data.

\begin{figure}[t]
	\centering
	\includegraphics[width=\linewidth]{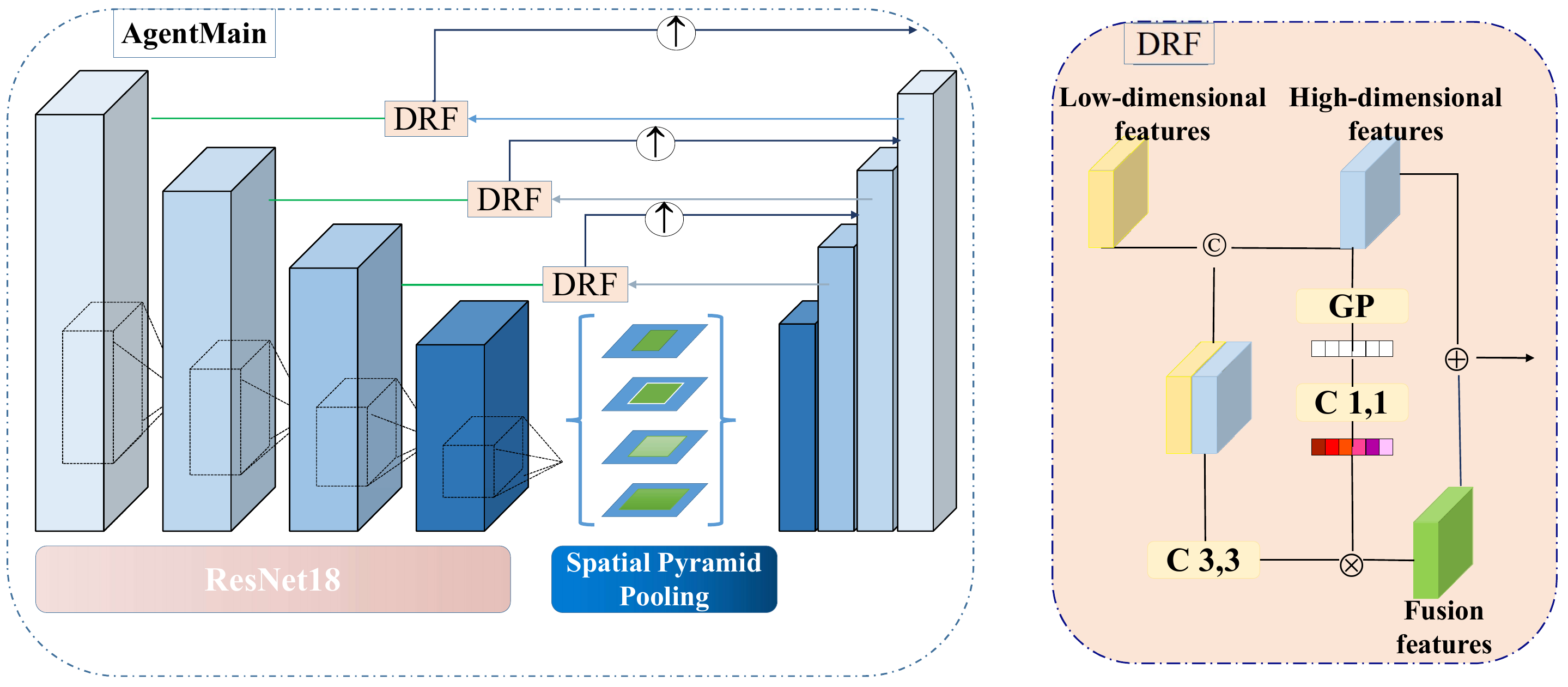}
	\caption{The  structure of Main agent.}
	\label{Fig3}
%\vspace{-8px}
\end{figure}

\textbf{Main agent.} As shown in Fig.~\ref{Fig3}, our main agent uses the classic encoder-decoder structure, which has achieved remarkable results in semantic segmentation tasks. The backbone of the encoder uses ResNet18~\cite{he2016deep}, and we use the spatial pyramid pool proposed in Deeplabv3+~\cite{chen2017deeplab} to enhance the features of the encoder output.

It is worth mentioning that, to use high-dimensional semantic information more effectively, we innovatively propose a dynamic residual feature fusion module (DRF), which considers the use of residual structure to maintain category semantic information.
As shown in Fig.~\ref{Fig3}, our DRF first concatenates low-dimensional information with high-dimensional semantic information. Then uses a $3 \times 3$ convolutional layer to reduce some channels, and finally, merges high-dimensional information with low-dimensional information.
We use deep separable convolution~\cite {chollet2017xception} to make the model more efficient.
To extract effective weights from high-dimensional features, we use global pooling to process the feature channels first, and then use two batched $1 \times 1$ convolutional layers to process the results, and finally use sigmoid for activation.

\begin{figure}[t]
	\centering
	\includegraphics[width=\linewidth]{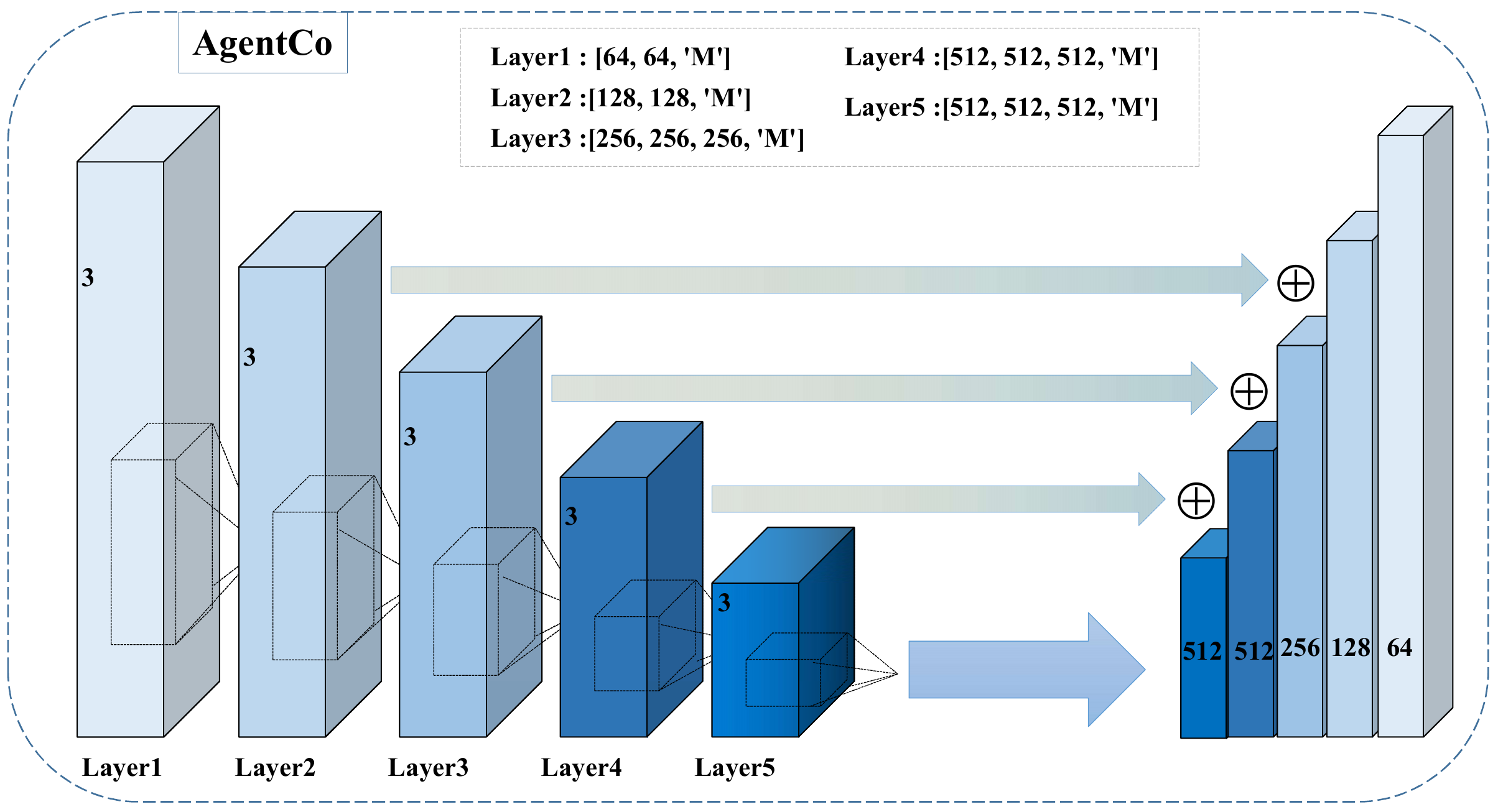}
	\caption{The  structure of cooperative agent.}
	\label{Fig4}
%\vspace{-8px}
\end{figure}

\textbf{Collaborative agent.} As shown in Fig.~\ref{Fig4},  our collaborative agent uses the FCN8~\cite{long2015fully} model, which uses a fully convolutional neural network, and the encoder uses VGG16~\cite{Simonyan15} as the backbone.
We use two trained agents to deal with unlabeled data, and then each agent will have different prediction results.
The DLA task is essentially a pixel-level classification, the pixel classification results of the two prediction results.
We will select a part of the data whose difference is more significant than 25\% and mark it manually. This comparison difference is an empirical value.

\subsection{Agent Update}
The purpose of agent update is to allow the agent to learn unknown knowledge from new samples. Currently, it is mostly achieved by fine-tuning the model. We designed four fine-tuning strategies for this and compared them. The first is to directly use key samples to fine-tune the model, the second considers the use of key samples and initial samples for mixed fine-tuning, and the third considers the use of key samples and initial sample re-tuning.  The fourth considers the distribution gap between the new data and the original data.
We follow the idea of reinforcement learning and assign new weights to selected samples and basic data (Eq.~\ref{equ:update4})  to make full use of the data.
As shown in Eq.~\ref{equ:update4}, ${{\lambda _1}}$ and ${{\lambda _2}}$ are two hyperparameters that directly act on the loss function.
By the experiment, we have determined that the value of ${{\lambda _1}}$ is 0.2 and the value of ${{\lambda _2}}$ is 0.8. This model can achieve the best effect.
We will further prove our update method's effect in section ~\ref{sec:ABstudy}.

\section{Experiments}
\label{sec:exp}
%\subsection{Metric}
 Following the setting of the paper~\cite{yang2017learning,li2020docbank, sarkar2020document, sarkhel2021improving}, we evaluate our model using accuracy, precision, recall and F1 as metrics.

\vspace{0.08in}
\noindent\textbf{Metric.} Several metrics are used to evaluate the performance.
We first define $M$ as the $n\times n$ confusion matrix with $n$ categories.
\emph{Accuracy} (\emph{Acc}) is the ratio of the pixels that are correctly predicted in a given image, \ie,
\begin{equation}
Acc=\frac{{\sum\nolimits_i {{M_{ii}}} }}{{\sum\nolimits_{ij} {{M_{ij}}} }}
\label{equ:Acurracy}
\end{equation}
\emph{Precision} (\emph{P}) is the ratio that is actually a positive example in the example that is divided into positive examples, \ie,
\begin{equation}
\begin{aligned}
P = \frac{1}{n}\sum\limits_{i = 1}^n {{P_i}} \qquad {P_i} = \frac{{{M_{ii}}}}{{\sum\nolimits_j {{M_{ji}}} }}
\end{aligned}
\label{equ:Percision}
\end{equation}
\emph{Recall} (\emph{R}) measures the coverage. There are multiple positive examples of metrics that are divided into positive examples, \ie,
\begin{equation}
\begin{aligned}
R = \frac{1}{n}\sum\limits_{i = 1}^n {{R_i}}  \qquad {R_i} = \frac{{{M_{ii}}}}{{\sum\nolimits_j {{M_{ij}}} }}
\end{aligned}
\label{equ:Recall}
\end{equation}
$F_1$ is an indicator used to measure the accuracy of a binary model. It also takes into account the accuracy and recall rate of the classification model. The $F_1$ score can be seen as a weighted average of model accuracy and recall:
\begin{equation}
{F_1} = \frac{{2 \cdot P \cdot R}}{{P + R}}
\label{equ:F1}
\end{equation}

\subsection{Datasets}
\noindent
\textbf{DSSE-200.} The DSSE-200~\cite{yang2017learning} is a comprehensive dataset, which comes from magazine screenshots, PPT screenshots, book cover screenshots, and old newspapers. It contains 200 images, including seven categories.

\noindent
\textbf{CS-150.} Clark ~\etal~proposed  the CS-150~\cite{clark2015looking}, it consists of 150 papers and includes 1175 images.
The classification criteria in this dataset are divided into three types: image, table, and others.

\subsection{Implementation Details}
\textbf{DSSE-200.}
First, we train AgentCo and AgentMain by using synthetic samples.
We split the training set and validation set into 8: 2.
Because the generated sample layout is not complicated enough, to prevent over-fitting, we constrain F1 below 95\%.
Then we use AgentCo and AgentMain to deal with the image of DSSE-200 and obtain pixel classification results of all images, and then we compare the prediction results of the two agents for the same image, find some images with an error rate of more than 25\%.
For the DSSE-200, we found 23 images, we manually mark these images.
Besides, we extracted the materials in these images, and then randomly selected some resources from the resource library, finally put them into our document synthesis tool to generate some samples.
Finally, we synthesize 77 images and use the new 100 images to continue to retrain the agent.

\noindent
\textbf{CS-150.}
The CS-150 is relatively simple, and it is composed of simple paper layouts. Therefore, the result is better processed.
We limit the samples with a different rate of more than 5\% on the first page to obtain 9 images.
These 9 images were manually marked.
Besides, we extracted the materials in these images, and then randomly selected some resources from the resource library, finally put them into our document synthesis tool to generate some samples.
Finally, we synthesize 91 images and use the new 100 images to continue training the model.

% \usepackage{booktabs}
% \usepackage{multirow}
\begin{table}[t]
\centering
\caption{F1 scores (\%) for page segmentation on the datasets. The Baseline means a migration model that is not trained on the new dataset.}
\begin{tabular}{@{}p{45px}p{12px}<{\centering}p{12px}<{\centering}p{12px}<{\centering}p{16px}<{\centering}|p{12px}<{\centering}p{12px}<{\centering}p{12px}<{\centering}p{12px}<{\centering}@{}}
\toprule
\multirow{2}{*}{Method} & \multicolumn{4}{c}{DSSE-200} & \multicolumn{4}{|c}{CS-150} \\ \cmidrule(l){2-9}
                        & A     & P     & R     & F1   & A     & P    & R    & F1   \\ \cmidrule(r){1-9}
Baseline                & 80.6  & 77.9  & 76.3  & 77.1 & 92.9  & 87.6 & 88.5 & 88.0 \\
KSS                     & 86.2  & 85.1  & 87.5  & 86.3 & 98.5  & 96.2 & 95.1 & 95.6 \\ \bottomrule
\end{tabular}
\label{tabel1}
\end{table}

\begin{figure}[t]
	\centering
	\includegraphics[width=1\linewidth]{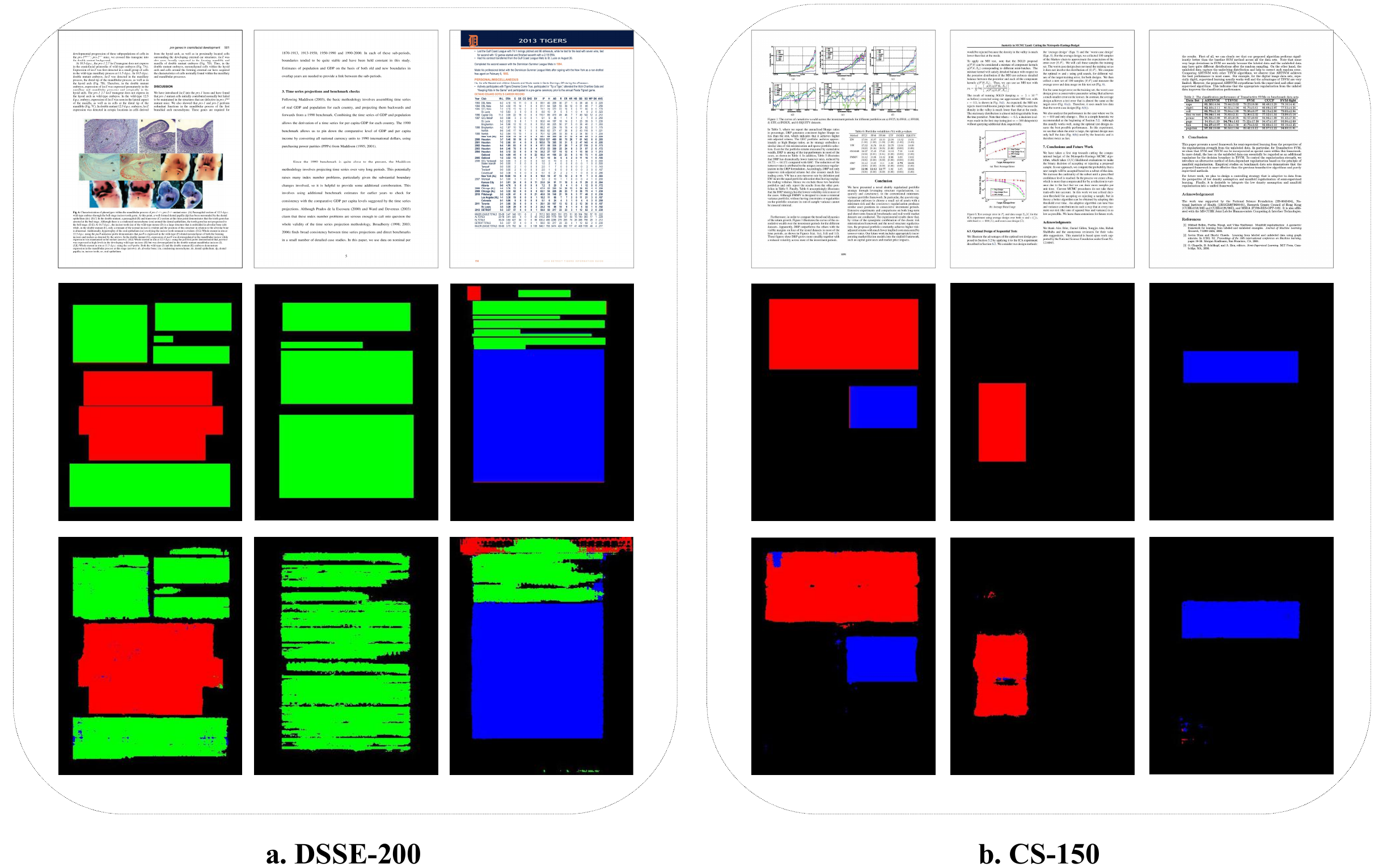}
	\caption{Example real documents and their corresponding segmentation of DSSE-200 and CS-150. Top: original. Middle: ground-truth. Bottom:predictions. Segmentation label colors are: \colorbox{red}{figure}, \colorbox{blue}{\color{white}{table}}, \colorbox{yellow}{text} and \colorbox{black}{\color{white}{background}} (The CS-150 mark uses text as background).}
	\label{FigE}
%\vspace{-8px}
\end{figure}

\subsection{Results and Comparisons}
\noindent
\textbf{DSSE-200.}
To get fair results, we remove the key samples as a test set. So, the DSSE-200 test set contains 177 images.
We obtain the results as shown in Table~\ref{tabel1}, and we deal with the visualization results as shown in Fig.~\ref{FigE}.
The baseline model trained with generated data.
As shown in Table~\ref{tabel1}, we used 23 labeled data to improve the performance by 9.2\%, so we can see that our method has a certain effect, and we will compare it with randomly selected samples later.

\vspace{2px}
\noindent
\textbf{CS-150.}
The results we got are shown in Table~\ref{tabel1}, and the visualization results drawn are shown in Fig.~\ref{FigE}.
As shown in Table~\ref{tabel1}, we used 23 types of labeled data to increase the performance of the datasets by 7.6\%.

\subsection{Comparisons with State-ot-the-arts}
%\textbf{DSSE-200}

\begin{table*}[t]
	\center
	\caption{F1 scores (\%) for page segmentation on the CS-150 dataset compare with state-of-the-arts.}
	\label{table: CPC}
	%\begin{tabular}{@{}p{76px}p{16px}<{\centering}p{16px}<{\centering}p{20px}<{\centering}|p{16px}<{\centering}p{16px}<{\centering}p{16px}<{\centering}@{}}
	\begin{tabular}{lcccccc}
	\toprule
	\multirow{2}{*}{Method}                          & \multicolumn{3}{c}{figure}              & \multicolumn{3}{|c}{table}                                \\ \cmidrule(l){2-7}
													 & $P$                & $R$                & $F_1$              & $P$       & $R$      & $F1$         \\ \cmidrule(l){1-7}
	
	PDFPlots~\cite{praczyk2013automatic}             & 0.624            & 0.500            & 0.555           & 0.429            & 0.363           & 0.393  \\
	PDFFigures~\cite{clark2015looking}               & 0.961            & 0.911            & 0.935           & 0.962            & 0.921           & 0.941  \\
	PDFFigures 2.0~\cite{clark2016pdffigures}        & 0.980            & 0.961            & 0.970           & \textbf{0.979}   & 0.963           & 0.971  \\
	KSS                                              & \textbf{0.984}             & \textbf{0.971}        & \textbf{0.977}            & 0.978           & \textbf{0.972 }  & \textbf{0.974}      \\ \bottomrule
	\end{tabular}
	\vspace{8px}
	\end{table*}

Due to the differences in page classification granularity requirements, DLA tasks also have large differences in page elements' division.
We selected the previous two representative classification methods and compared them.
Following the setting of the paper~\cite{yang2017learning}, we have an experiment on the DSSE-200.
The experiment shows the following Table~\ref{table: CPA} and Table~\ref{table: CPA1}.
As can be seen from Table~\ref{table: CPA} and Table~\ref{table: CPA1}, whether it is fine-grained classification or coarse-grained classification, our method is the most effective.
Following the setting of the paper~\cite{clark2016pdffigures}, we test and compare the CS-150.
The results are shown in Table ~\ref{table: CPC}.
Observing the experimental results in Table~\ref{table: CPC}, our results reached the state-of-the-art.
\begin{table}[t]
\centering
\small
\caption{IoU scores (\%) for fine-grained page segmentation on the DSSE-200 dataset compare with state-of-the-arts. BG means background.}
\label{table: CPA}
\begin{tabular}{@{}p{30px}p{16px}<{\centering}p{16px}<{\centering}p{16px}<{\centering}p{18px}<{\centering}p{18px}<{\centering}p{16px}<{\centering}p{15px}<{\centering}p{15px}<{\centering}@{}}
%\begin{tabular}{@{}lcccccccc@{}}
\toprule
Method                         & BG    & figure          & table          & section           & caption          & list           & para.         & mean            \\ \midrule
MFCN~\cite{yang2017learning}   & \textbf{83.9}          & 83.7            & 79.7           & 59.4              & 61.1    & 68.4           &79.3  & 73.3             \\
KSS                            &82.2    &\textbf{84.9}    & \textbf{82.6}  &    \textbf{72.5}             & \textbf{67.5}             &   \textbf{75.1}           &   \textbf{84.3}          & \textbf{78.3}    \\ \bottomrule
\end{tabular}
\vspace{8px}
\end{table}

% \usepackage{booktabs}
\begin{table}[t]
\center
\caption{IoU scores (\%) for coarse-grained page segmentation on the DSSE-200 dataset compare with state-of-the-arts.}
\label{table: CPA1}
\begin{tabular}{@{}p{95px}p{58px}<{\centering}p{58px}<{\centering}@{}}
\toprule
Methods                                        & no-text & text                     \\ \midrule
Leptonica ~\cite{bloomberg2007document}        & 84.7    & \multicolumn{1}{c}{86.8} \\
Bukhari~\etal~\cite{bukhari2011improved}       & 90.6    & 90.3                     \\
MFCN (binary)~\cite{yang2017learning}          & 94.5    & 91.0                     \\
KSS  (binary)                                  & \textbf{ 96.4 }         &  \textbf{ 93.2 }                         \\ \midrule \midrule
Methods                                        & figure  & text                     \\ \midrule
Fernandez~\etal~\cite{fernandez2012document}   & 70.1    & 85.8                     \\
MFCN (binary)~\cite{yang2017learning}          & 77.1    & 91.0                     \\
KSS  (binary)                                  & \textbf{ 82.6 }         & \textbf{ 91.2}                          \\ \bottomrule
\end{tabular}
\vspace{8px}
\end{table}

In order to get a fair comparison, we recode several current more competitive structures and compared them with our proposed structure.
Because the DSSE-200 and CS-150 do not provide standard training datasets, we randomly select some images as the train data and others as test data.
We split DSSE-200 and CS-150 into the training set and test set according to 6:4.
We compared the model performance on the DSSE-200 and CS-150 datasets using the conventional training mode, and the results are shown in Table~\ref{tabel4}.
It can be seen from Table~\ref{tabel4} that KSS has outstanding performance.

\begin{table}[t]
\center
\caption{F1 scores (\%) for page segmentation on dataset. The Para represents the trainable parameter amount of the model, the 'M' means million.}
\label{tabel4}
\begin{tabular}{@{}p{45px}p{10px}<{\centering}p{10px}<{\centering}p{10px}<{\centering}p{15px}<{\centering}|p{10px}<{\centering}p{10px}<{\centering}p{10px}<{\centering}p{10px}<{\centering}p{10px}<{\centering}c@{}}
\toprule
\multicolumn{1}{c}{\multirow{2}{*}{Method}}                  & \multicolumn{4}{c}{DSSE-200} & \multicolumn{4}{|c}{CS-150} & \multicolumn{1}{c}{{Para}}     \\ \cmidrule(l){2-9}
 \multicolumn{1}{c}{}                                        & A     & P     & R     & F1   & A     & P    & R    & F1   &  \multicolumn{1}{c}{(M)}   \\  \midrule
Segnet~\cite{badrinarayanan2017segnet}   & 83.6  & 81.8  & 82.6  & 82.2 & 99.3  & 93.6 & 93.9 & 93.7 & 29  \\
PSPnet~\cite{zhao2017pyramid}            & 84.6  &  \textbf{83.5}  & 84.1  & 83.8 & 99.4  & 93.9 & 93.3 & 93.6 & 46  \\
PANet~\cite{li2018pyramid}               & 81.1  &82.6  & 85.2  & 83.9 & 99.2  & 93.7 & 94.0 & 93.8 & 168 \\
KSS                                      &  \textbf{85.9}  & 82.5  &  \textbf{86.3}  &  \textbf{84.4} &  \textbf{99.4}  &  \textbf{94.5} &  \textbf{94.7} &  \textbf{94.6} & 15  \\ \bottomrule
\end{tabular}
%\vspace{-8px}
\end{table}

\subsection{Ablation Study}
\label{sec:ABstudy}

\begin{figure*}
	\centering
	\includegraphics[width=0.92\linewidth]{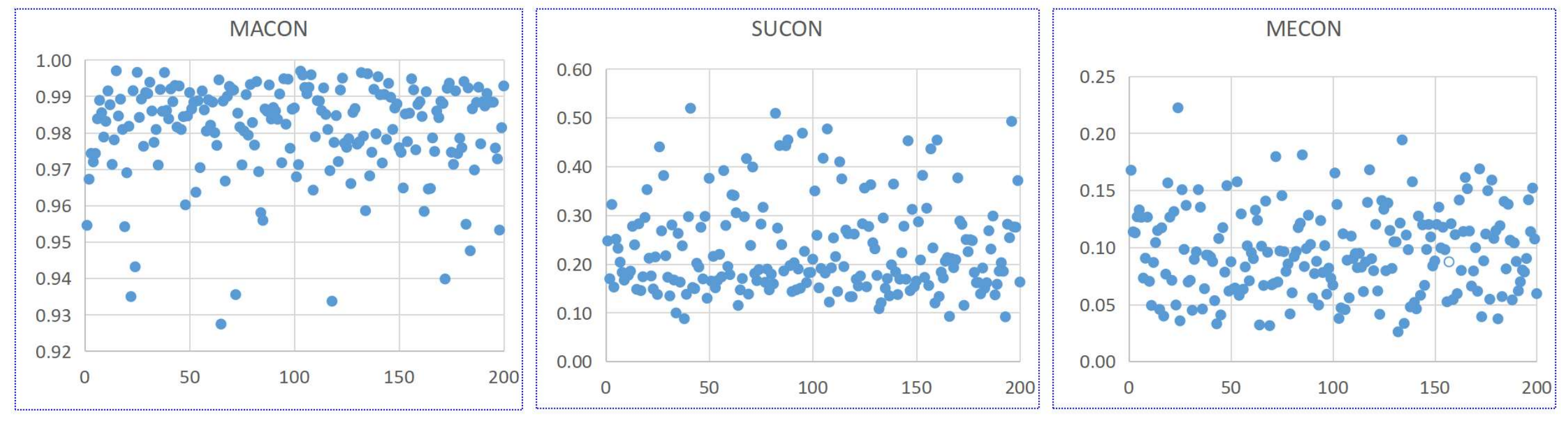}
	\caption{For different methods to select sample confidence results, MACON can refer to Eq.~\ref{equ:con1}, SUCON can refer to Eq.~\ref{equ:con2}, and MECON can refer to Eq.~\ref{equ:con3}.}
	\label{FigE3}
%\vspace{-3px}
\end{figure*}

\vspace{2px}
\noindent
\textbf{Validation of Agent Collaboration Mode}
To verify the agent collaboration mode, we considered four modes.
The first is to use two agents of the same structure (AgentMain+ AgentMain) and use the same datasets for training (MMOD);
the second is to use two agents of the same structure (AgentMain+ AgentMain) and use two different datasets for training (MMTD);
the third is to use two agents of different structures (AgentMain+ AgentCo) and use the same datasets for training (MCOD);
the fourth is to use two agents of different structures (AgentMain+ AgentCo) and use two different datasets for training (MCTD).
We directly use the selected samples to fine-tune the network, and the results are shown in Table~\ref{tabel5}.
It can be seen from the results in Table~\ref{tabel5} that the synergy effect of using two different structure models is better than that of using the same structure model. Using the same dataset to train the model to select key examples seems more important for agent update.

\begin{table}[t]
	\centering
	\caption{F1 scores (\%) for page segmentation on the DSSE-200. SN: The Number of sample.}
\begin{tabular}{@{}p{40px}p{30px}<{\centering}p{30px}<{\centering}p{30px}<{\centering}p{30px}<{\centering}p{30px}<{\centering}@{}}
\toprule
Method        & A    & P    & R    & F1 &SN  \\ \midrule
MMOD      & 68.6 & 66.4 & 69.8 & 68.1 & 19  \\
MMTD      & 77.6 & 73.9 & 71.0 & 72.4 & 59 \\
MCOD      & 78.1 & 79.2 & 77.9 & 78.5 & 23 \\
MCTD      & 75.2 & 76.2 & 71.7 & 73.9 & 37 \\ \bottomrule

\end{tabular}
\label{tabel5}
%\vspace{-8px}
\end{table}

\vspace{2px}
\noindent
\textbf{Validation of Key Samples Selection}
The previous classic methods used confidence as a measure of sample selection to deal with simple classification comparisons.
DLA is a pixel-level classification task.
We follow the work of Wan~\etal~\cite{wan2020human} and replace the classification confidence with the new confidence.
We choose the Three different confidence (Eq.~\ref{equ:con1}, Eq.~\ref{equ:con2}, Eq.~\ref{equ:con3}) of the entire picture as the new confidence.
The experimental results show as Fig.~\ref{FigE3} that the confidence interval is (0.92, 1], (0, 0.6), and (0, 0.25)).
As we can see in Fig.~\ref{FigE}, there are local texts on the table or in the figure, so the confidence method is not applicable.
We conducted experiments on DSSE-200 to compare the advantages of the sample selection strategy.
We split DSSE-200 into the training set and test set according to 6:4. First, we use the model that trains by synthesis data to test directly on the test set (Baseline). Then, we randomly select \emph{x} samples from the training set and use them fine-tuning models (Baseline+RD\emph{x}), and finally, we use KSS to select the key samples (Find 18 key samples) and use them fine-tuning models. And the results are shown in Table~\ref{tabel3}.
Obviously, compared with other methods, the KSS not only achieves better results but also uses very few training samples.

\begin{table}[t]
	\center
	\caption{F1 scores (\%) for page segmentation on DSSE-200 dataset by ablation study.}
\begin{tabular}{@{}p{80px}p{30px}<{\centering}p{30px}<{\centering}p{30px}<{\centering}p{30px}<{\centering}@{}}
\toprule
Method        & A    & P    & R    & F1   \\ \midrule
Baseline           & 79.9 & 79.3  & 75.2  & 77.2 \\
Baseline+RD20      & 81.3 & 80.2 & 81.5 & 80.8 \\
Baseline+RD40      & 82.9 & 81.6 & 83.2 & 82.4 \\
Baseline+RD60      & 84.5 & 84.7 & 86.9 & 85.8 \\
KSS                & 85.6 & 86.2 & 86.3 & 86.2 \\ \bottomrule
\end{tabular}
\label{tabel3}
%\vspace{-8px}
\end{table}

\vspace{2px}
\noindent
\textbf{Validation of Agent Update}
In order to compare the effects of our model update mode, we compared four model update methods.
First, we use a simple fine-tuning mode, and our learning rate is generally reduced to 1/10 of the initial training (PSFT).
We then use basic data and selected data for new fine-tuning and use the previous pre-trained model (PBSFT).
Next, we use a model initialized only with random numbers for training and used basic data and selected data(RBSFT).
The last one is the update method used by our framework (RBSRE).

It can be seen from the table ~\ref{tabel9} that only using key samples for fine-tuning results is not good.
Because the number of key samples is limited, only using key samples for fine-tuning will push the model to overfit.
Using a pre-training model can improve the performance of the model because the pre-training model contains more basic knowledge and can better improve the generalization of the model.
The RBSRE can obtain better results, RBSRE not only considered key samples but also focus on the base samples.

\begin{table}[t]
	\center
	\caption{F1 scores (\%) for page segmentation of Agent Update on the DSSE-200.}
\begin{tabular}{@{}p{80px}p{30px}<{\centering}p{30px}<{\centering}p{30px}<{\centering}p{30px}<{\centering}@{}}
\toprule
Method        & A    & P    & R    & F1   \\ \midrule
PSFT (Eq.~\ref{equ:update1})     & 68.9 & 70.1 & 68.2 & 69.1 \\
PBSFT (Eq.~\ref{equ:update2})      & 82.7 & 81.6 & 82.5 & 82.0 \\
RBSFT (Eq.~\ref{equ:update3})      & 78.6 & 77.5 & 79.3 & 78.4 \\
RBSRE (Eq.~\ref{equ:update4})      & 86.2 & 85.1 & 87.5 & 86.3 \\ \bottomrule
\end{tabular}
\label{tabel9}
%\vspace{-8px}
\end{table}

\section{Conclusions}
\label{sec:conclusion}

In this paper, we proposed a key sample selection framework based on HITL named KSS.
KSS includes three components: agent initialization, agent collaboration, and agent update.
The agent initialization uses LaTex as a general sample generation engine to let the agent learn more common knowledge.
We have innovatively proposed the multi-agent collaboration that realizes the selection of key samples.
Hence, we revisited the learning system from reinforcement learning and designed a sample-based agent update strategy, which effectively improves the agent's ability to accept new samples.
To the best of our knowledge, this is the first work to introduce agent collaboration into HITL.
Our results indicate that we have improved the state of the art on previously established benchmarks, and KSS can better find key samples by using fewer data.
\\\\
\noindent\textbf{Acknowledgment.}
This work was supported in part by the 2020 East China Normal University Outstanding Doctoral Students Academic Innovation Ability Improvement Project (YBNLTS2020-042), the Science and Technology Commission of Shanghai Municipality (19511120200).

\bibliographystyle{IEEEtran}
\bibliography{total}

\end{document}